\pdfoutput=1

\documentclass[11pt]{article}

\usepackage[final]{acl}

\usepackage{times}
\usepackage{latexsym}
\usepackage[T1]{fontenc}

\usepackage[utf8]{inputenc}

\usepackage{microtype}

\usepackage{inconsolata}

\usepackage{graphicx}
\usepackage{booktabs}
\usepackage[normalem]{ulem}
\useunder{\uline}{\ul}{}
\usepackage{enumitem}
\usepackage{float}
\usepackage{kotex}
\usepackage{subcaption}
\usepackage{listings}
\title{A Practical Approach for Building Production-Grade Conversational Agents with Workflow Graphs}

\author{
  Chiwan Park\thanks{Equal contribution}\;\;
   Wonjun Jang\footnotemark[1]\;\;
   Daeryong Kim\footnotemark[1]\;\;
   Aelim Ahn\;\;
   Kichang Yang\;\;
\\
   \textbf{Woosung Hwang\;\;
   Jihyeon Roh\;\;
   Hyerin Park\;\;
    Hyosun Wang\;\;} \\
  \textbf{Min Seok Kim\thanks{Corresponding authors}\thanks{Contact: {\{marko.k, don.kang\}@kakaocorp.com}}\;\;
  Jihoon Kang\footnotemark[2]\footnotemark[3]\;\;}
  \\\\
  Kakao
}

\begin{document}
\maketitle

\begin{abstract}
The advancement of Large Language Models (LLMs) has led to significant improvements in various service domains, including search, recommendation, and chatbot applications.
However, applying state-of-the-art (SOTA) research to industrial settings presents challenges, as it requires maintaining flexible conversational abilities while also strictly complying with service-specific constraints.
This can be seen as two conflicting requirements due to the probabilistic nature of LLMs.
In this paper, we propose our approach to addressing this challenge and detail the strategies we employed to overcome their inherent limitations in real-world applications.
We conduct a practical case study of a conversational agent designed for the e-commerce domain, detailing our implementation workflow and optimizations.
Our findings provide insights into bridging the gap between academic research and real-world application, introducing a framework for developing scalable, controllable, and reliable AI-driven agents.
\end{abstract}
\section{Introduction}
\label{sec:introduction}

Large Language Models~\citep{chatgpt,gpt4,claude,llama} have exhibited exceptional performance improvement across various language tasks, making them highly valuable in numerous industries.
Beyond their language task performance, several works~\cite{toolformer, react,toolllm} demonstrate the model's ability to effectively utilize external tools to tackle complex tasks in various domains, including coding~\cite{codeagent}, travel planning~\cite{travelplanner}, recommendation~\cite{recmind}, and scientific research~\cite{aicoscientist}.
This ability rapidly led to the advancements of Conversational Agents, which aim to assist users with real-world tasks, such as booking restaurants or purchasing gifts, by interacting with external systems.

Despite their excellent performance, many challenges still exist in building real-world agents~\cite{rschprdgap}.
First, because of the nature of the probabilistic next-token generation of LLMs, the agents randomly fail to comply with business requirements for specific domains.
For example, considering a conversational e-commerce agent, the agent should retrieve the exact metadata of products to prohibit recommending cigarettes or alcohol to an underage user.
However, occasionally, the agent uses its pre-trained knowledge instead of retrieving the external metadata, resulting in a wrong hallucinated response~\cite{toolbehonest}.
This drawback becomes particularly apparent in cases where strict compliance with business requirements exists.
Second, there is a general demand for response formatting capabilities for the agent.
In the case of mobile-targeted agents, due to their small screen size, the model should respond with a specific format, such as a length limit and emoji bullets.
Furthermore, for certain products, the e-commerce agent must strictly comply with specific constraints, such as avoiding hype or exaggerated advertisements or ensuring proper attribution and source citation.
Last, prompt engineering involves writing detailed descriptions into the system prompt to ensure that LLMs follow these requirements.
The more detailed requirements are, the longer the system prompts will be;
thus, the comprehensive system prompt degrades the latency and accuracy of response~\cite{sametask-moretkns}.

\begin{figure}[tbh]
    \centering
    \includegraphics[width=\columnwidth]{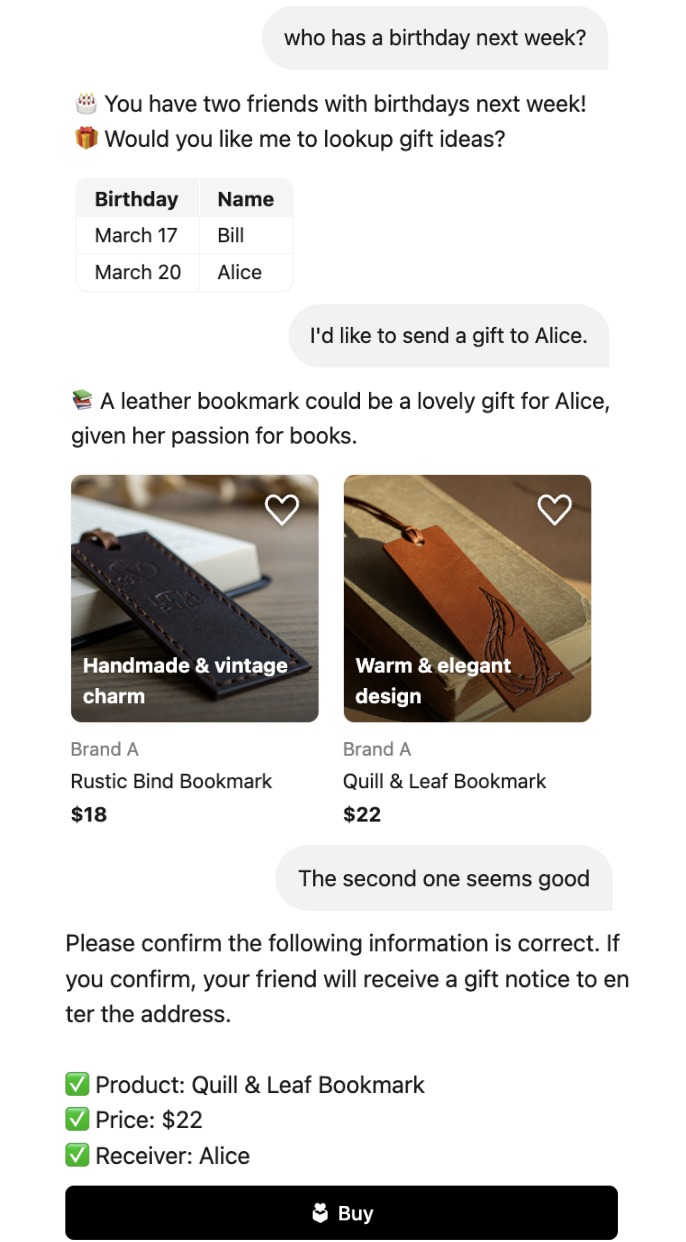}
    \caption{A mobile messenger conversation between a user and our e-commerce agent. The first two turns require external tool calls to respond without hallucination. There are also output format constraints to make the responses readable in a mobile environment, such as emoji bullets.}
    \label{fig:chat-example}
\end{figure}

In this paper, we present our work on building a conversational e-commerce agent that runs on a mobile messenger.
Figure~\ref{fig:chat-example} shows an example conversation between a user and the agent about purchasing a birthday gift for a friend.
The agent helps the user explore products through search and recommendation, obtain detailed information about a product, and purchase the product.
Furthermore, using social relationships in the messenger platform, the user can check the birthdays of his/her friends and send gifts.
We adopt a hybrid approach that leverages a directed acyclic graph (DAG) workflow to guide the agent’s behavior, instead of relying on end-to-end generation from LLMs. 
This design enables flexible interactions while ensuring strict compliance with scenario-specific requirements.
While the DAG framework efficiently handles the complex business requirements, fine-tuning becomes nontrivial since each message of the chat history comes from different states. 
To tackle this problem, we present a dataset construction and training approach that enables effective fine-tuning despite state-dependent chat histories.
We begin with converting our requirements into a workflow graph.
Then, we implement the workflow as a prototype agent with LLMs and several system prompts.
After gathering annotated conversations between human annotators and the prototype agent, we used the conversations to train our agent models carefully.
We repeat this process iteratively to achieve the required response quality.
Thanks to the hybrid approach and training, the agent shows a 52\% improvement in task accuracy and a 50\% improvement in format adherence compared to the baseline, outperforming GPT-4o performance.
Our main contributions are as follows:

\begin{itemize}
    \item \textbf{Multi-State DAG Framework:} Real-world agents must comply with many scenario-dependent constraints.
    We present a graph-based framework, each state with distinct prompts, tools and execution rules adhering to the specific constraints of the state. Traversing the graph seamlessly represents the wide range of expected scenarios, while efficiently distributing constraint handling across appropriate states.
    \item \textbf{Training Strategy on DAG Framework:} We introduce a dataset construction and training strategy specifically designed to overcome the challenges posed by state-dependent message histories in our DAG framework. This further enhances the precision of the agent to meet even the stringent demands of sensitive domains such as e-commerce.
    \item \textbf{Real-World Example:} We show a real-world working example using the two methods above. Our empirical results clearly demonstrate that even state-of-the-art LLMs fall short in achieving satisfactory performance in the e-commerce domain, underscoring the necessity of our proposed hybrid approach for practical deployment.
\end{itemize}

\section{Background}
\label{sec:background}

\subsection{Conversational Agents}
\label{sec:conversational-agents}

Traditional dialog-based frameworks such as Rasa~\cite{rasa} and Talkamatic~\cite{larsson2016talkamatic} manage conversations using rule-based state tracking, offering reliability and interpretability. However, they often lack the flexibility and reasoning capabilities of modern LLM-based agents.

Recent advances in LLMs such as GPT-4~\cite{gpt4}, Claude~\cite{claude}, Mixtral~\cite{mixtral}, Qwen~\cite{qwen25}, and Deepseek~\cite{deepseekv3} have driven rapid progress and shifted expectations regarding the fluency and capabilities of conversational agents.
LLMs can invoke external tools when provided with natural-language descriptions and instructions.
Toolformer~\cite{toolformer} demonstrates how LLMs formulate calls of external tools with appropriate parameters based on a few examples and textual instructions.
ToolLLM~\cite{toolllm} shows that LLMs can use multiple external tools to answer user questions.
Agents with reasoning capabilities like ReAct~\cite{react}, Chain-of-Thoughts~\cite{cot}, and Tree-of-Thoughts~\cite{tot} show significant performance improvements.

As LLM-based agents become more capable and widely adopted, much effort has also been devoted to evaluating their performance across various domains, such as general agents~\cite{agentbench,agentboard}, travel planning~\cite{travelplanner}, games~\cite{gamebench}, coding~\cite{codeagent}, and scientific research~\cite{aicoscientist}.
These evaluation methods vary slightly in detail, but they all essentially measure how successfully a requested task has been accomplished.

\subsection{Challenges in Production-grade Conversational Agents}
\label{sec:challenges}

Even with recent advances in LLM-based agents, significant challenges still remain in building production-grade conversational agents~\cite{healthcareagent,rschprdgap,multiagent}.
One major limitation of existing approaches is their narrow focus on the task accuracy of agents' execution results. This overlooks several crucial aspects, including specific requirement following and output formatting, which can be equally important in assessing an agent's performance regarding production-grade agents~\cite{trustagent}.
For example, consider an e-commerce agent that recommends products to a user. If a recommendation includes a compliance-violating description, it should be regarded as a failure, even if the user ends up selecting the product.
Addressing such issues often requires more detailed and restrictive system prompts, which in turn increase inference costs due to longer context lengths.

Several industry-specific agent frameworks highlight the importance of such aspects.
For instance, Amazon Bedrock\footnote{\url{https://aws.amazon.com/bedrock/agents/}} offers post-processing steps to control the agent response.
Google Vertex AI Agents\footnote{\url{https://cloud.google.com/products/agent-builder}} adopts LangChain\footnote{\url{https://www.langchain.com/}}, an open-source framework for building agents with predefined workflows, to enhance adherence to requirements.
MARCO~\cite{marco} is a notable approach that considers not only the accuracy but also the validity of output formatting.
However, MARCO relies on a separate guardrail component to verify and retry faulty outputs using reflection prompts, which can significantly degrade both response latency and overall accuracy.

\subsection{Graph-based Agent Frameworks}
\label{sec:graph-based-agent}

Due to their high expressivity and controllability, graphs are widely adopted to model complex workflows in various agent frameworks, such as Dify\footnote{\url{https://dify.ai/}} and LangGraph\footnote{\url{https://langchain-ai.github.io/langgraph/}}.
In these frameworks, an agent $\mathcal{A}$ is modeled as a workflow graph $\mathcal{G}$, defined by a tuple $(\mathcal{V}, \mathcal{E})$ where $\mathcal{V}$ is a set of nodes and $\mathcal{E} \subset \mathcal{V} \times \mathcal{V}$ is a set of directed edges.
Each node $v \in \mathcal{V}$ has a computational routine $f_v$ that executes external tools for the agent, or LLMs.
The routine returns a tuple $(o_v, v_n)$ where $o_v$ is a tool or LLMs response, and $v_n$ is a successor node, one of the nodes connected to $v$ in graph $\mathcal{G}$.
From this graph structure, running the agent is considered as a graph traversal.
The agent starts with the initial node $v_{init}$, which is the entry node of the graph. It iteratively moves to the successor nodes until it reaches the final node $v_{final}$ and returns its output $o_{v_{final}}$.

While existing frameworks simplify the construction and deployment of graph-based LLM agents, our research focuses on methodologies for achieving production-grade responses, including a practical approach to fine-tuning tightly coupled graph–LLM agents.

\section{Methodology}
\label{sec:methodology}

\begin{figure*}[t]
    \centering
    \includegraphics[width=\textwidth]{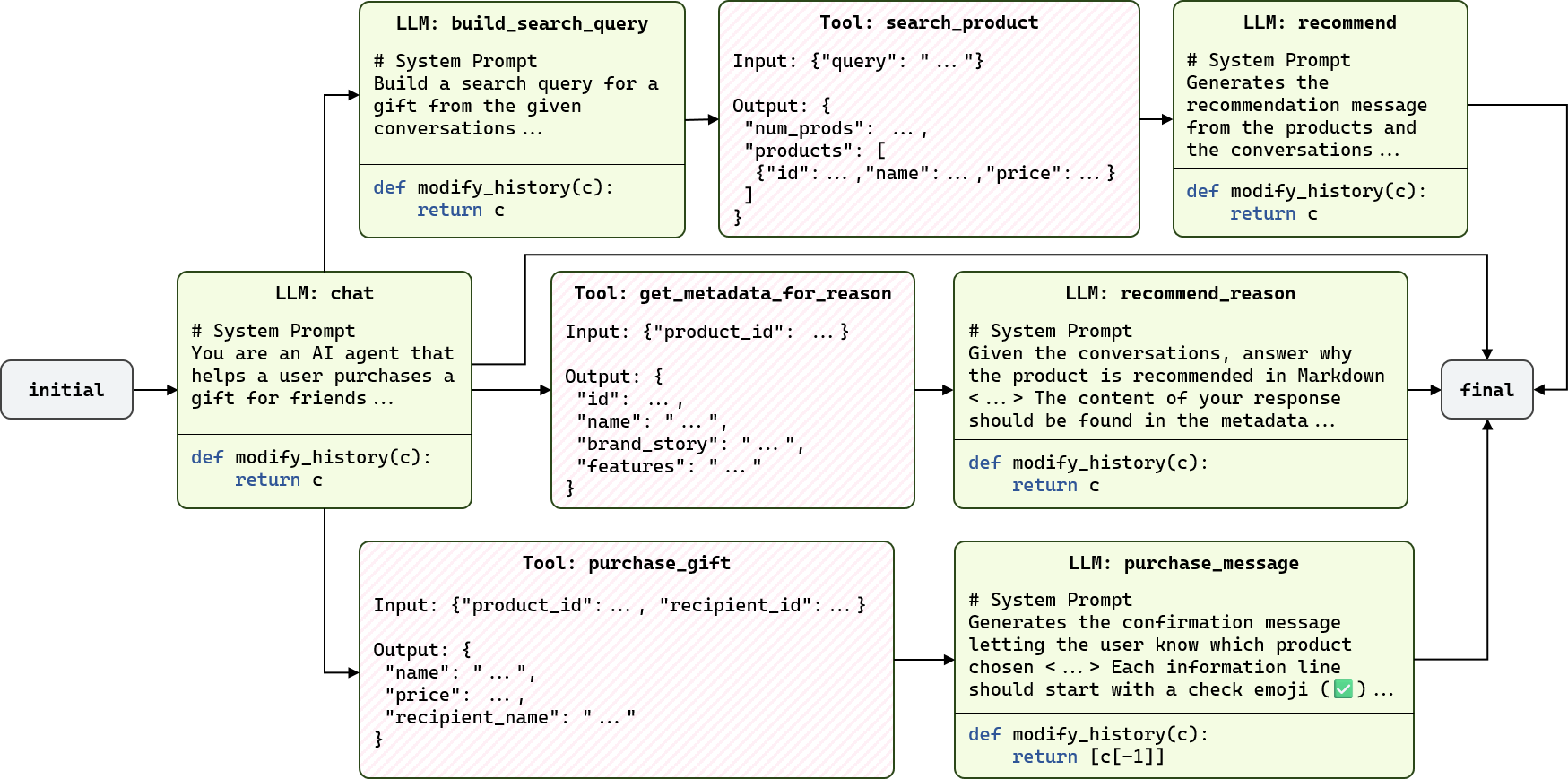}
    \caption{An example workflow graph. Each LLM calling node (green colored) has its system prompt and a custom routine (\texttt{modify\_history}) to manipulate conversation histories. The tool nodes (pink striped) are used to call pre-defined external tools and have the schemas for input and output. For clarity, we only show nodes related to gift recommendations and omit some content of the system prompts, including few-shot examples of the responses.}
    \label{fig:workflow-graph}
\end{figure*}

In this section, we introduce our framework to build conversational agents with an example of an e-commerce agent.
We convert our agent workflow into a workflow graph, build a prototype agent with a general LLM to collect a high-quality dataset, and train LLMs to enhance the agent's behavioral control in complex tasks.

\subsection{Workflow Graph Design}
\label{sec:workflow-graph}

While following compliance is crucial to building production-grade conversational agents, we observe that LLMs often struggle to adhere to complex conditional rules.
We add a specific structure to nodes calling LLM in the workflow graph to enhance compliance-following stability.
Each LLM calling node $v$ has its system prompt $s_v$ describing constraints applied in that context with few-shot examples and a custom computational routine that manipulates the given conversational history to prevent the LLM from hallucinating.
Each node that calls an external tool has two schemas: one for input and another for output.
Constrained decoding~\cite{outlines,xgrammar} is applied if the output from the LLM node needs to be passed as input to the tool node.

Figure~\ref{fig:workflow-graph} illustrates an example workflow graph for our agent.
Nodes that invoke LLMs are shown in green, while those that call external tools are depicted with pink diagonal stripes.
Each LLM-invoking node is associated with a system prompt that encodes rules such as Markdown formatting and emoji usage (e.g., \texttt{recommend\_reason}, \texttt{purchase\_message}).
By default, these nodes use the full conversation history as input.
However, some nodes, like \texttt{purchase\_message}, remove all the previous conversation history except the purchase information (the last turn of history) by their \texttt{modify\_history} subroutine.
This manipulation helps mitigate hallucination by limiting access to irrelevant prior information.

In the example, you can see that the workflow graph is designed with a general-purpose chat node (\texttt{chat}) as its initial entry point. From the node, the LLM may suggest tool calls to effectively route to appropriate task-specific nodes, or handle out-of-scenario user queries via general chat. In the latter case, it may respond using its internal knowledge or gently guide the user toward a more relevant task. 
After reaching \texttt{final} you can restart from \texttt{initial} for multi-turn conversations.

\subsection{Data Collection with Prototype Agent}
\label{sec:data-collection}

We constructed a dataset comprising conversations between human annotators and our agent, structured as a list of $(x_i, o_i)$ pairs, where $x_i$ denotes the $i$-th input message from an annotator and $o_i$ is the corresponding agent response.
The data collection process consists of three steps: (1) building a prototype agent, (2) recording interactions, and (3) correcting erroneous examples.

\paragraph{Building a Prototype Agent} A key challenge in collecting data for agents handling complex tasks is generating appropriate agent responses.  
Annotators can easily answer simple questions, such as "Who are you?", but struggle with queries that involve multi-step reasoning, such as "Recommend a wine that goes well with sirloin steak."  
This difficulty arises because they must consider and imagine multiple steps, including tool calls and workflow graph traversals, to generate a single answer.
To address this, we built a prototype agent using GPT-4o and our workflow graph to generate initial draft responses that annotators could then refine.

\paragraph{Recording Interactions} In this step, annotators interact with the prototype agent as end users.  
The agent automatically records all interactions, including the full graph traversal history and the results of any external tool calls.

\paragraph{Correcting Erroneous Examples} The final step involves reviewing and correcting erroneous agent responses.  
Annotators examine all interactions and outputs for each conversation and revise any errors they identify.  
To assist with this process and reduce human error, we provide automated checkers that help detect issues and verify corrections.  
One particularly useful tool is a static type checker for tool call arguments, which are typically structured as JSON objects.  
Annotators often produce ill-formatted JSON, especially when dealing with complex schemas.

\subsection{Fine-Tuning with Response Masking}
\label{sec:fine-tuning}

We employ a fine-tuning approach with the dataset to enhance the agent's stability.
For each node $v$ calling LLMs, we formulate the agent interactions into a chatbot-style sequence $(s_v, x_1, o_1, x_2, o_2, ..., x_n, o_n)$ where $s_v$ is a system prompt for the node $v$, $x_i$ denotes $i$-th observations (user messages or tool results), and $o_i$ is $i$-th response of the agent.

Standard multi-turn training strategies often optimize the model on all assistant outputs in the conversation history. 
However, in the graph-based agent setting, this can degrade the model's ability to follow system prompts consistently, as responses in the same conversation may originate from different nodes with distinct instructions.

For example, consider a workflow graph with two LLM nodes $v_1$ and $v_2$, a conversation history for $v_1$ can be formulated as $(s_{v_1}, x_1, o_1, x_2, o_2, x_3, o_3)$ where $o_2$ is generated by $v_2$, while the other responses $o_1$ and $o_3$ are generated by $v_1$.
In such a case, training on $o_2$ under the prompt $s_{v_1}$ would introduce conflicting supervision, as $o_2$ reflects the constraints of $v_2$.

To address this, we apply loss masking during training, excluding responses generated by other nodes from the loss calculation.  
This prevents the model from learning under mismatched prompt constraints and helps maintain system prompt fidelity for each node.

\section{Experiment}
\label{sec:experiment}

\begin{table*}[h]
\centering
\small
\begin{tabular}{@{}r|rrr|rrr|rrr|rr@{}}
\toprule
\multicolumn{1}{l|}{} & \multicolumn{3}{c|}{\textbf{Qwen 2.5 (32B)}} & \multicolumn{3}{c|}{\textbf{Gemma 3 (27B)}} & \multicolumn{3}{c|}{\textbf{Internal Model}} & \multicolumn{2}{c}{\textbf{GPT-4o}} \\ \midrule
\textbf{Metric} & \multicolumn{1}{c}{\textbf{B}} & \multicolumn{1}{c}{\textbf{WG}} & \multicolumn{1}{c|}{\textbf{WG-FT}} & \multicolumn{1}{c}{\textbf{B}} & \multicolumn{1}{c}{\textbf{WG}} & \multicolumn{1}{c|}{\textbf{WG-FT}} & \multicolumn{1}{c}{\textbf{B}} & \multicolumn{1}{c}{\textbf{WG}} & \multicolumn{1}{c|}{\textbf{WG-FT}} & \multicolumn{1}{c}{\textbf{B}} & \multicolumn{1}{c}{\textbf{WG}} \\ \midrule
Accuracy & 0.578 & 0.616 & 0.884 & 0.622 & 0.711 & 0.887 & 0.744 & 0.790 & \textbf{0.890} & 0.864 & {\ul 0.888} \\ \midrule
Format Adherence & 0.734 & 0.813 & {\ul 0.969} & 0.692 & 0.882 & 0.966 & 0.655 & 0.951 & \textbf{0.987} & 0.778 & 0.964 \\ \midrule
Response Validity & 2.816 & 2.831 & 2.880 & 2.821 & 2.849 & {\ul 2.911} & 2.893 & 2.874 & \textbf{2.953} & 2.856 & 2.882 \\ \bottomrule
\end{tabular}
\caption{Qualitative results on our test dataset. The accuracy and format adherence are the ratio of valid responses over the total, while the response quality is rated between 1 and 3. For each model, we evaluate multiple agent architectures including Basic (B), Workflow Graph (WG), and Workflow Graph with Fine-Tuning (WG-FT). The top performance of each metric is marked as bold, and the second one is underlined.}
\label{tbl:result}
\end{table*}

In this section, we detail our experiments to evaluate agents in our service scenarios.



\subsection{Experimental Setting}
\label{sec:experimental-setting}

\paragraph{Dataset}
We used a subset of our dataset collected as described in Section~\ref{sec:data-collection}.  
The test set contains 161 conversations between the human annotators and the agent, containing 2100 turns.

\paragraph{Evaluation Protocol}
We conducted turn-level assessments following previous studies~\cite{autoagents,worfbench}.
Each turn is paired with a reference response annotated by the annotators, and evaluated across three dimensions:
First, we measure accuracy, which indicates whether the agent selects the correct tool and provides appropriate arguments.  
Due to the flexibility of certain arguments (e.g., search queries for gift recommendations), we employ an LLM-as-a-Judge approach~\cite{llm-as-a-judge} to verify argument validity.
Second, we assess format adherence, which checks whether the agent's response conforms to the predefined message format using a strictly coded validator.
Finally, we evaluate response quality using the LLM-as-a-Judge method, comparing the agent's response to the reference in terms of clarity, helpfulness, and relevance.
The first two metrics are binary (0 or 1), while response quality is scored on a 3-point scale (1 to 3).

\paragraph{Model}
We evaluated both open-source and proprietary LLMs to show that our approach is general for various models and is not limited to our internal model.
We use Qwen 2.5 32B~\cite{qwen25} and Gemma 3 27B~\cite{gemma3} for open-source baselines as their model sizes are comparable to our internal model and align well with our performance and latency goals.
Our internal model also falls within the 27B-32B parameter range. It is built upon an open-source base model and further trained on internal datasets to better support Korean, the target service language, more details are provided in Appendix~\ref{appendix:internal}.
For proprietary LLMs, we use GPT-4o\footnote{The specific model version is gpt-4o-2024-11-20.}, one of the strongest SOTA models currently available and presumably larger in scale.
We use it as a high-end baseline to provide a performance reference point for our experiments.
We use o3-mini\footnote{\url{https://openai.com/index/openai-o3-mini/}} as a judge for the LLM-as-a-Judge evaluation, leveraging its reasoning ability to judge with complex rules.

\paragraph{Agent Architecture}
We tested four agent architectures.
Basic (B) is a baseline architecture that uses a single system prompt and a tool-calling mechanism proposed by the original model providers. 
In this setting, we concatenate all node-specific instructions\textemdash such as compliance constraints and output formatting rules\textemdash into a single prompt without structural separation.
Workflow Graph (WG) is our workflow graph-based architecture, as we describe in Section~\ref{sec:workflow-graph}.
Workflow Graph with Fine-Tuning (WG-FT) is an agent with a fine-tuned model by the method described in Section~\ref{sec:fine-tuning}.

\subsection{Results}

Table~\ref{tbl:result} summarizes the experimental results of our agents compared to the baselines.  
Due to its strong general performance, GPT-4o achieves the highest score for all metrics among the models for the basic agent architecture.
However, GPT-4o still fails to consistently follow the required output formatting.
Other open-source models, such as Gemma 3 27B and Qwen 2.5 32B, also suffer from incorrect tool selection and low accuracy.

Applying our workflow graph structure to the agents enhances format adherence and accuracy for all models.
The accuracy is improved by up to 14\% over the basic architecture.
Formatting errors are dramatically reduced thanks to the shorter and more focused system prompts in our workflow graph.
For our internal model, the format adherence improved from 0.655 to 0.951, representing a 45\% relative improvement.
The format adherence of other models also increased by up to 27\%.
Response quality also improved for most models under the graph-based architecture, with only a negligible drop observed for the internal model.

The fine-tuning with response masking further improves the agent in all metrics, making our internal model-based agent outperform the GPT-4o-based one.
Other open-source models also achieve comparable performance with GPT-4o across all evaluation metrics.

\subsection{Human Assessment}

\begin{table}[t]
\centering
\small
\begin{tabular}{@{}r|r@{}}
\toprule
\textbf{Evaluation} & \multicolumn{1}{l}{\textbf{Internal \textgreater{}= GPT-4o (\%)}} \\ \midrule
Regular chat & 42.42 \\ \midrule
Safety & 60.53 \\ \midrule
Product recommendation & 82.42 \\ \midrule
Messenger-related features & 60.61 \\ \midrule
\textbf{Overall} & 63.29 \\ \bottomrule
\end{tabular}
\caption{Human assessment results on our e-commerce agent in a real-world environment. The testers are provided with two responses from our model and GPT-4o, and they are requested to choose better models.}
\label{tbl:human}
\end{table}

We deployed \textit{AI Shopping Mate}\footnote{\url{https://mate.kakao.com/shopping}} on both Kakao\-Talk\footnote{\url{https://www.kakaocorp.com/page/service/service/KakaoTalk?lang=en}} application and the web. (see Appendix~D for details).
The agent covers over one million products across various categories.
In this real-world setting, we conducted comparative "battle" tests similar to Chatbot Arena~\cite{chatbotarena}, evaluating our internal model against GPT-4o.  
All external systems and integrations connected to the agent were kept identical across both models.  
Each tester submitted a message and received two anonymized responses\textemdash one from each model.  
They were then asked to select the better response or mark them as a tie.
Table~\ref{tbl:human} summarizes the results of the human assessment.  
We categorize the requests into four types:  
(1) Regular chat,  
(2) Safety\textemdash requests intended to provoke unsafe or inappropriate outputs,  
(3) Product recommendation, and  
(4) Messenger-related features such as birthday reminders.

Our agent using the internal model outperformed the GPT-4o-based agent in all categories except regular chat.  
From follow-up interviews, we found that language fluency significantly influenced human preference\textemdash an aspect that was difficult to capture via LLM-as-a-Judge evaluation.  
We leave further investigation of this aspect for future work.

\section{Conclusion}
\label{sec:conclusion}

In this study, we presented our framework for building conversational agents that address key challenges in utilizing LLMs and graphs for complex and necessary compliances.
We demonstrated that our agent with the internal model outperforms the GPT-4o-based agent for our e-commerce agent scenarios.
Our framework's generic design allows it to be adapted for agents across various domains wherever complex tasks need to be executed correctly.
\section{Limitations}
\label{sec:limitations}

Our framework has several limitations in terms of data collection and evaluation.  
First, the data collection process is highly human-dependent, requiring significant time and effort from annotators.  
Moreover, the collected conversations may exhibit demographic bias, as the annotator pool was limited in terms of gender and age.  
As a potential remedy, LLM-based simulation where an LLM acts as a user interacting with the agent could be explored in future work.

Second, evaluating response quality remains a challenge. 
Although we define rules for high-quality responses and employ LLM-as-a-Judge with reference answers, this approach may not fully reflect human preferences.  
To further support the validity of our evaluation framework, future work could examine the correlation between human judgments and LLM-based assessments more systematically.

\section*{Ethical Considerations}

In this work, we incorporate multiple safeguard mechanisms to ensure the safe and ethical use of our conversational agent.  
Real-time filtering is applied to both user inputs and model outputs to mitigate hate speech, stereotyping, and sensitive social content.  
A multi-layered policy distinguishes between generalized group criticism and statements based on personal experience, guiding the model to maintain neutrality even in borderline cases.

To protect personal information and rights, our system detects sensitive data such as social security and bank account numbers in real time.  
It also issues warnings for content potentially related to intellectual property violations and enforces uniform responses when risks are detected.  
In addition, we apply annotation guidelines designed to minimize personal bias by differentiating between unjust generalizations and fact-based individual descriptions.

\section*{Acknowledgments}

We would like to thank the following team members and contributors for their valuable support throughout the development of the AI Shopping Mate service.  
We thank the AI Model Platform Development Team\textemdash including Oseok Han, Jeonghyeon Lee, Bomi Hong, Hyukjin Kwon, Junyoung Jeong, and Minho Gil\textemdash for their work on building the agent platform.  
We also thank the Adaptive AI Team\textemdash including Hyungsuk Noh, Songmin Han, Yongwook Jeong, Suin Lee, Taehyun Jung, Kyushik Min, and Gyuju Han\textemdash for constructing the service metadata.  
We are grateful to Geonhee Lee, Jongmyung Gong, and Hyunwoo Yoo for developing the search functionality.  
We also thank Yuri Lee and Jihye Park for their support in service collaboration, and Sooyeon Lee and Dain Kim for their contributions to service enhancement.  
Finally, we thank the annotators from Linkage Lab for their assistance in data collection.

\bibliography{references}

\clearpage
\appendix
\section*{Appendix}

\section{Implementation Details}

\paragraph{Training}
We implement our fine-tuning strategy using the Axolotl framework,\footnote{\url{https://github.com/axolotl-ai-cloud/axolotl}} which supports flexible dataset construction and various parameter-efficient fine-tuning methods such as LoRA~\cite{lora}.  
We adopt LoRA-based fine-tuning and optimize hyperparameters based on validation loss.
To apply our proposed loss masking method, we leverage Axolotl's support for segment-level input masking, allowing us to exclude responses generated by irrelevant nodes from the loss calculation.  
After fine-tuning, we merge the adapters into the base models to reduce latency during the serving phase.

\paragraph{Serving}
To serve our models, we use vLLM~\cite{kwon2023efficient} and SGLang~\cite{zheng2024sglangefficientexecutionstructured} to deploy open-weight and internal models.
We also build our custom agent platform traversing our workflow graph.
The platform is responsible for communicating with our models, executing external tools, and delivering responses to end users.

\section{Internal Model}
\label{appendix:internal}

We use an internal LLM in our experiments. While the model is not publicly disclosed, we provide details to support reproducibility. It is built upon an open-source model in the 27B-32B parameter range. To adapt it for Korean-language services, we conducted additional continuous pretraining and instruction tuning using internal Korean datasets.  
After the tuning stage, we applied model merging techniques~\cite{goddard-etal-2024-arcees} to refine performance across both general-purpose and domain-specific tasks.  
Our internal model serves as a key testbed in our experiments and is comparable in scale to Qwen 2.5 32B and Gemma 3 27B.

\section{Evaluation Prompts}
\label{sec:evaluation-prompts}

Figures~\ref{fig:tool-accuracy} and \ref{fig:response-quality} are prompts for evaluating the accuracy of tool execution and response quality using the LLM-as-a-Judge approach.
For each turn to be evaluated, we pack conversation history, tools, agent response, and a reference response in the same format as the prompt.
The judge LLM returns a score, which we extract from the output.  
If parsing fails, we retry until a valid score is obtained.

\begin{figure}[t]
    \centering
    \begin{lstlisting}
You are requested to evaluate the decided tool call by a language model. You are given the following information as follows:
  - <tools>: The list of tools that are available to the model.
    - <name>: The name of the tool.
    - <description>: The description of the tool.
    - <arguments>: The arguments that the tool receives.
  - <history>: The chat history between the user, the model and the tool response.
    - <message>: the message that was sent by the user, the model or the tool. The sender of the message is given as `role` attribute.
  - <reference_tool_call>: The reference answer that the model has decided to make.
    - <name>: The name of the tool.
    - <arguments>: The arguments that the tool will be called with.
  - <tool_call>: The tool call that the model has decided to make.
    - <name>: The name of the tool.
    - <arguments>: The arguments that the tool will be called with.

The tool call should be evaluated based on the following criteria:
  - The required arguments of the tool must be extracted.
  - The arguments should be extracted from the chat history.
  - If the tool requires some price or quantity ranges, they should be extracted from the chat history.
    - The start of the range should not be same as the end of the range.
  - The arguments extracted could be different from the reference tool call, but should be semantically similar.

Evaluate the arguments of tool call comparing it with the reference tool call, and determine whether the tool call is appropriate or not in terms of the criteria above.
Your response should be in the following format:
- Reason: <reason for the score in at most 3 sentences in one line>
- Score: <1 if the tool call is appropriate else 0>
\end{lstlisting}
    \caption{Evaluation Prompt for Task Accuracy.}
    \label{fig:tool-accuracy}
\end{figure}

\begin{figure}[t]
    \centering
    \begin{lstlisting}
You are requested to evaluate the linguistic quality of the generated response. You are given the following information as follows:
    - <history>: The chat history between the user, the model and the tool response.
      - <message>: the message that was sent by the user, the model or the tool. The sender of the message is given as `role` attribute.
    - <response>: The response generated by the model.
    - <reference>: The reference response that the model has respond.

  Evaluate the response based on the following criteria:
    - The content of response should match with that of the reference response.
    - The response should be written in Korean, unless there is a specific instruction to use another language.
    - The response should be fluent and natural.
    - The response should be grammatically correct.
    - The response MUST not contain unnecessary characters (such as Chinese characters, special characters, etc.) or non-understable characters. This is critical for the response to be considered valid.
    - The response should be completed, and contain no repeated or cut-off words.
    - The response will be presented in a small-size smartphone screen; thus, the following conditions should be also met.
      - All the tool results except `purchase_gift` tool results are displayed in the screen as cards. The duplicated response with the tool results should be considered as invalid.
      - Emoji-containing response is considered as good.

  Evaluate the response and score it on a scale of 1 to 3 in terms of the criteria above.
  - 1: not valid
  - 2: somewhat valid
  - 3: highly valid

  Your response should be formatted as follows:
  - Reason: <reason for the score in at most 3 sentences in one line>
  - Score: <score>

  Note that only the two lines in your response are allowed.
    \end{lstlisting}
    \caption{Evaluation Prompt for Response Quality.}
    \label{fig:response-quality}
\end{figure}

\section{Service Deployment}
We deployed \textit{AI Shopping Mate} into the Korean market in two forms: (1) as a chatbot in the KakaoTalk messenger and (2) as an independent web service.
Regardless of its form, our service provides the same features.
When users specify the gift context\textemdash recipient, occasion, and budget\textemdash the service delivers a personalized gift recommendation.
The service has been publicly available since December 2024 and is fully powered by the architecture described in this paper.
We are planning to integrate \textit{AI Shopping Mate} into KakaoTalk Gift~\footnote{\url{https://gift.kakao.com}}, a top-tier sending gift service with 20M users.

\begin{figure*}[htbp]
    \centering
    \begin{subfigure}[t]{0.25\paperwidth}
        \centering
        \includegraphics[width=0.25\paperwidth,keepaspectratio]{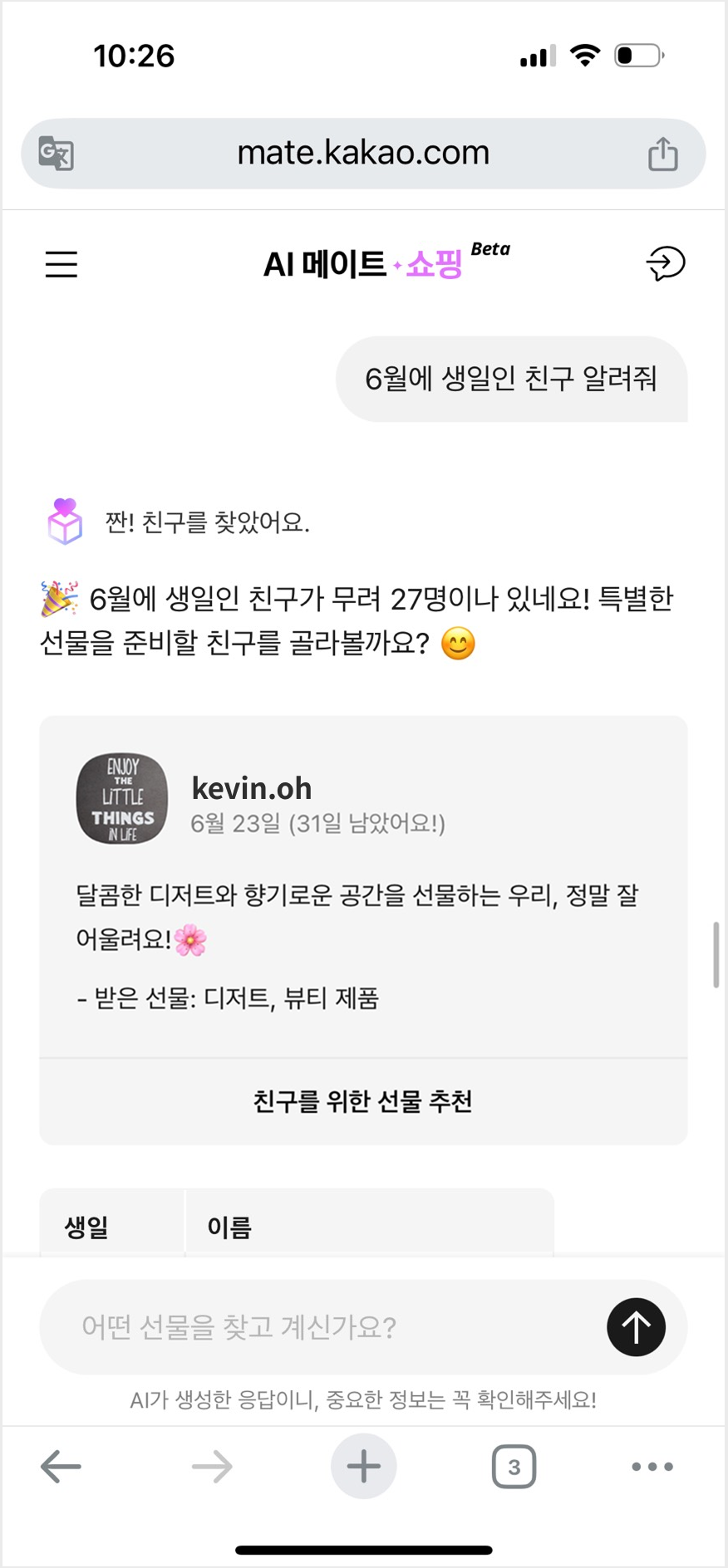}
        \caption{Finding friends to take care of}
        \label{subfig:friends}
    \end{subfigure}
    \hfill
    \begin{subfigure}[t]{0.25\paperwidth}
        \centering
        \includegraphics[width=0.25\paperwidth,keepaspectratio]{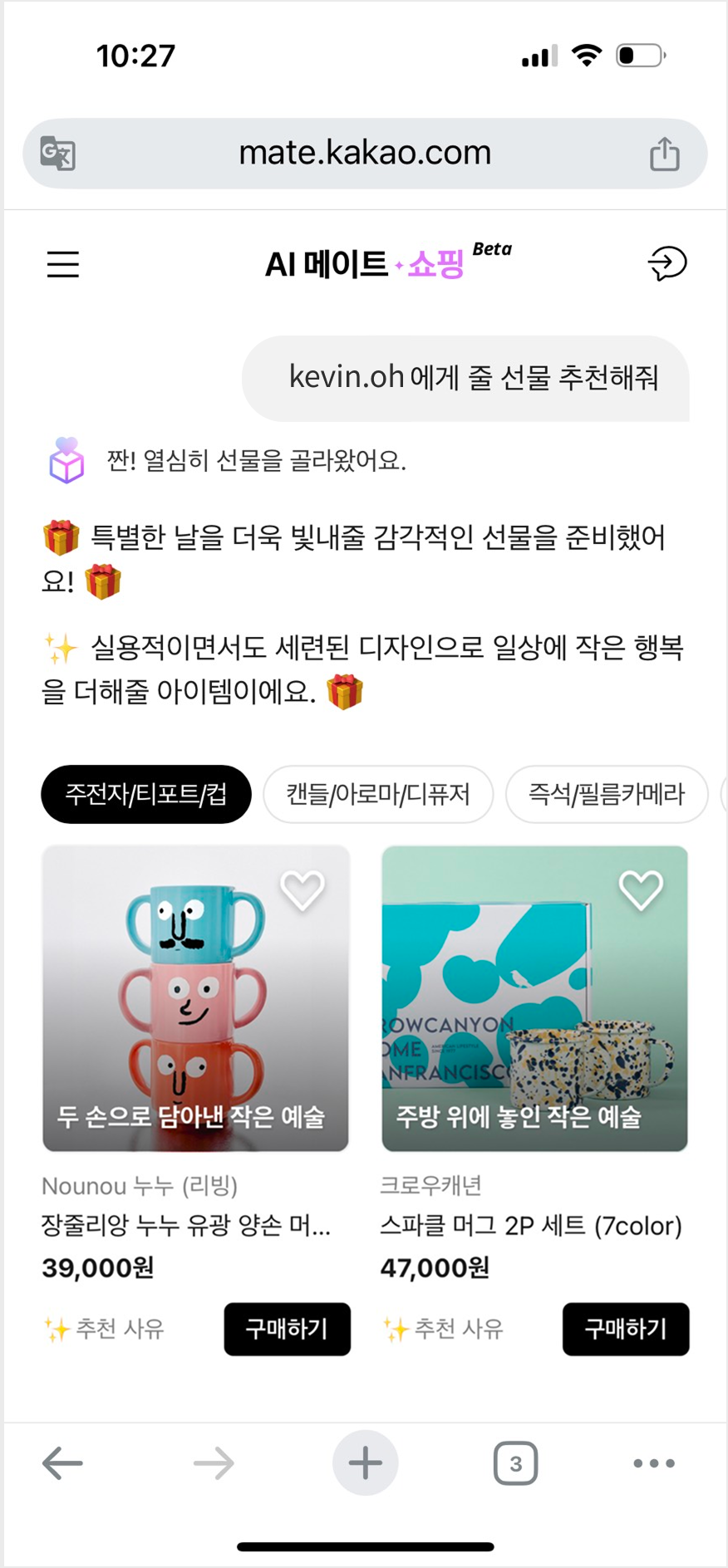}
        \caption{Recommending gifts for friends}
        \label{subfig:recommend-friends}
    \end{subfigure}
    \hfill
    \begin{subfigure}[t]{0.25\paperwidth}
        \centering
        \includegraphics[width=0.25\paperwidth,keepaspectratio]{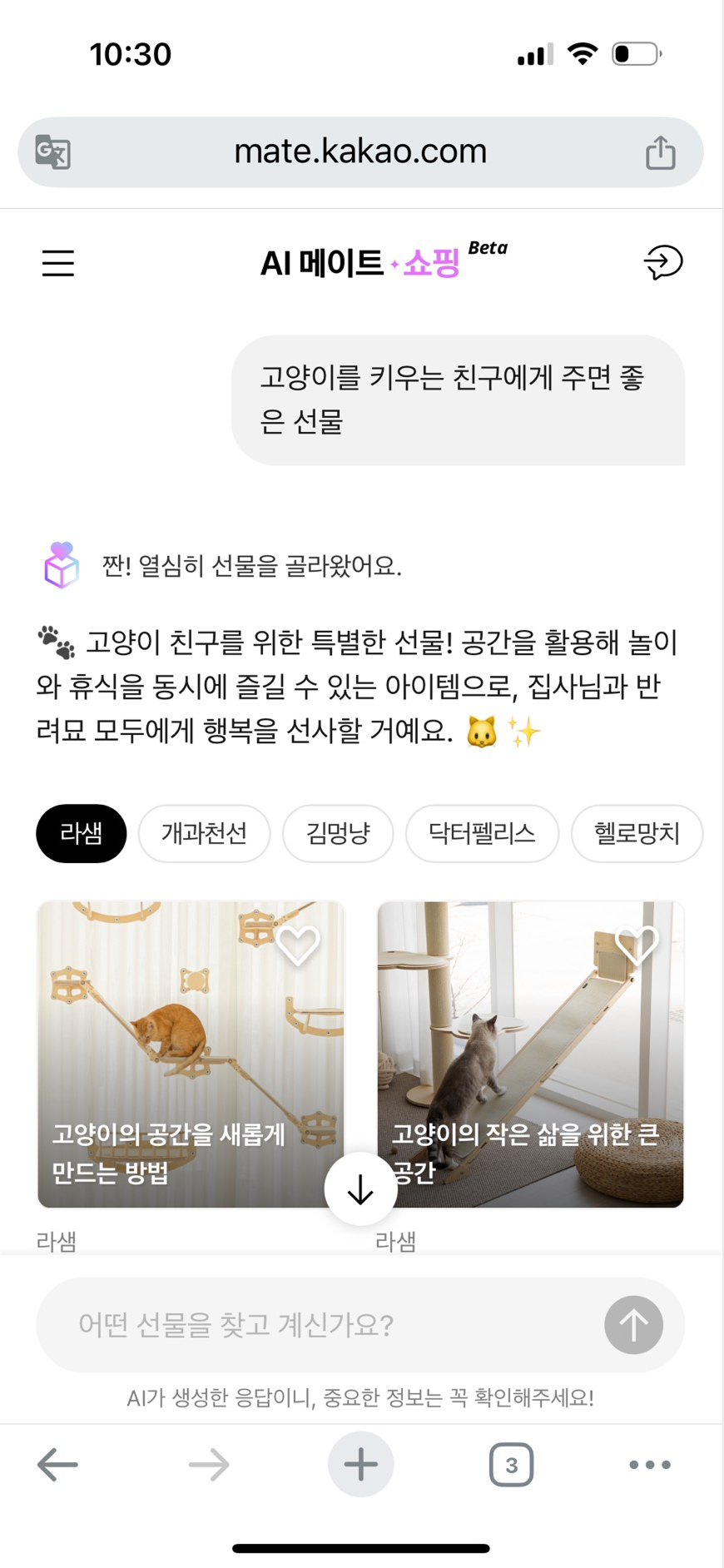}
        \caption{Recommending gifts from a context}
        \label{subfig:recommend-context}
    \end{subfigure}
    \vfill
    \begin{subfigure}[t]{0.55\paperwidth}
        \centering
        \includegraphics[width=0.25\paperwidth,keepaspectratio]{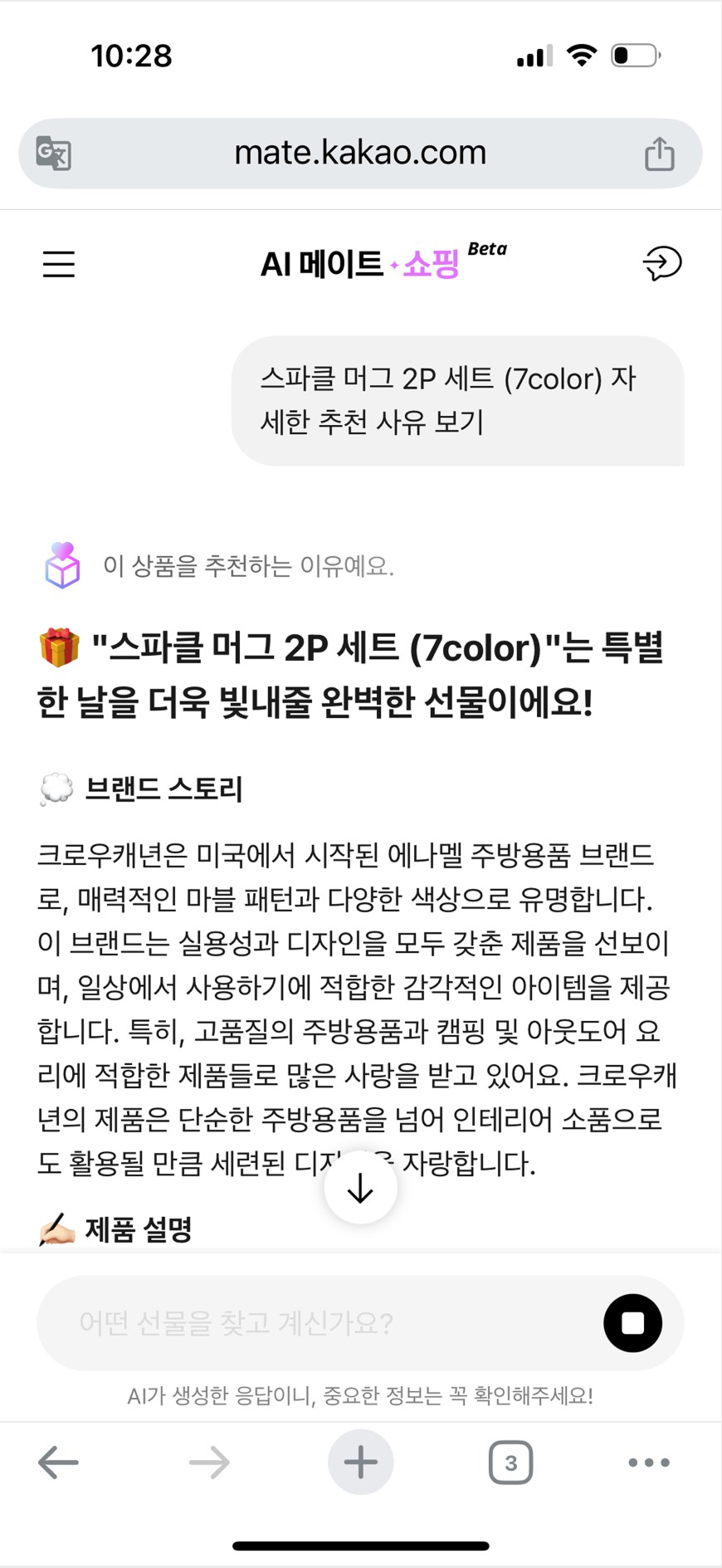}
        \includegraphics[width=0.25\paperwidth,keepaspectratio]{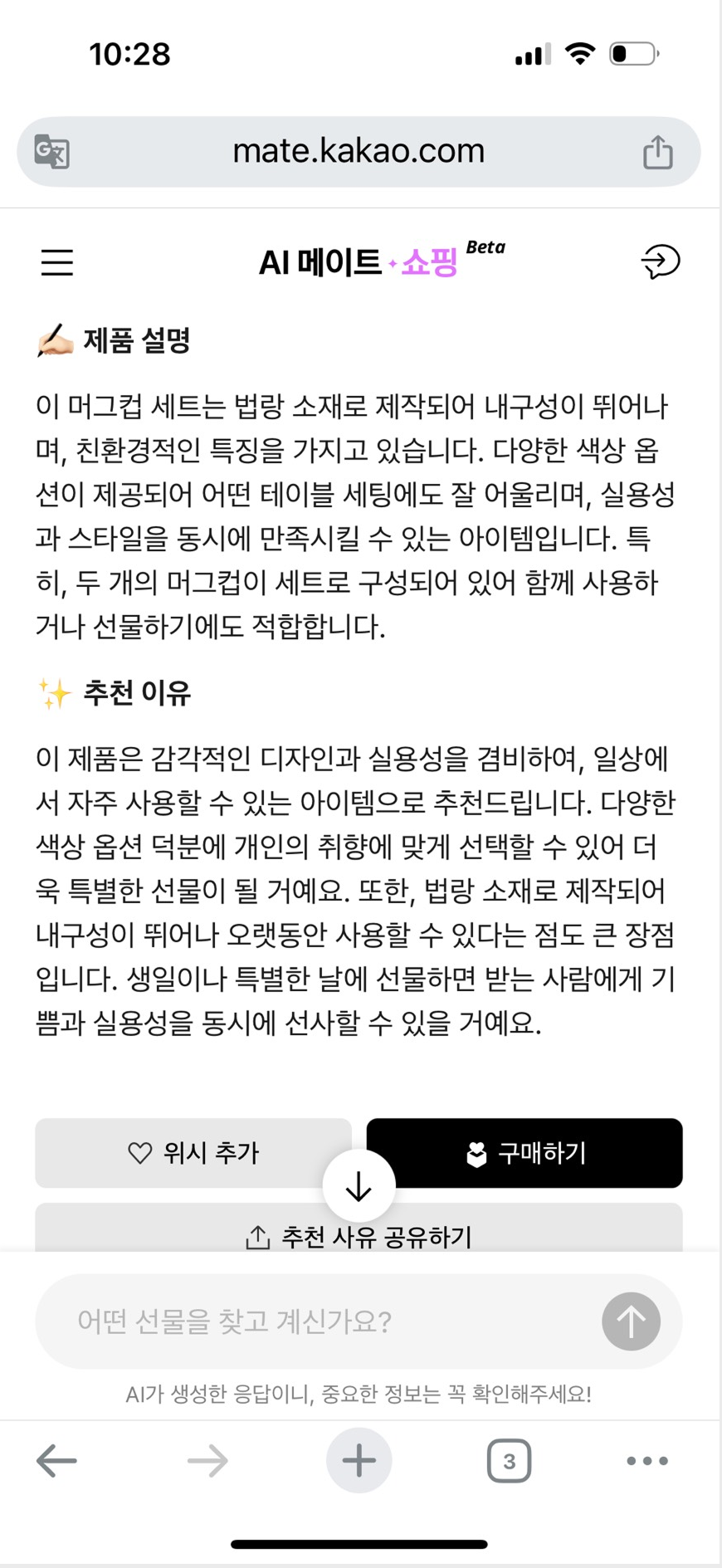}
        \caption{Providing detailed explanation for a recommended product}
        \label{subfig:reason}
    \end{subfigure}
    \caption{Example use cases on \textit{AI Shopping Mate}}
    \label{fig:aimate}
\end{figure*}

Figure~\ref{fig:aimate} presents example interfaces from the web-based version of our service.
Figure~\ref{subfig:friends} illustrates an instance where a user searches for friends whose birthdays fall in June.
In this scenario, the agent adheres to the specified response requirements, ensuring that each friend card displays the gifts previously exchanged with that friend.
Figures \ref{subfig:recommend-friends} and \ref{subfig:recommend-context} depict scenarios involving gift recommendations, either for a user's friend or based on the context from a user, respectively.
Figure~\ref{subfig:reason} demonstrates the provision of a detailed explanation for a recommended product.
It is noteworthy that, in accordance with our service's operational requirements, the agent first presents the brand story associated with the product before detailing the rationale for its recommendation.
Our workflow graph structure is adapted to meet these requirements.


\end{document}